\documentclass[letterpaper]{article}
\usepackage{aaai}
\usepackage{times}
\usepackage{helvet}
\usepackage{courier}
\usepackage{graphicx}

\usepackage{kotex}
\usepackage{xcolor}
\usepackage{ctable}

\usepackage{multirow}

\usepackage{figures/threeparttable}

 \graphicspath{ {figures/} }
\begin{document}
%

\title{Which Ads to Show? Advertisement Image Assessment\\with Auxiliary Information via Multi-step Modality Fusion}

\author{Kyung-Wha Park$^1$, JungHoon Lee$^2$, Sunyoung Kwon$^3$, Jung-Woo Ha$^3$, Kyung-Min Kim$^3$, Byoung-Tak Zhang$^1$$^,$$^4$$^,$$^5$ \\
$^1$ Interdisciplinary Program in Neuroscience, Seoul National University, $^2$ Statistics and Actuarial Science, Soongsil University,\\ 
$^3$ Clova AI Research, NAVER Corp, $^4$ Department of Computer Science and Engineering, Seoul National University, $^5$ Surromind Robotics\\
kwpark@bi.snu.ac.kr, ssutartup@gmail.com, \{sunny.kwon, jungwoo.ha, kyungmin.kim.ml\}@navercorp.com, btzhang@bi.snu.ac.kr\\
}

\maketitle
\begin{abstract}
Assessing aesthetic preference is a fundamental task related to human cognition. It can also contribute to various practical applications such as image creation for online advertisements. Despite crucial influences of image quality, auxiliary information of ad images such as tags and target subjects can also determine image preference. Existing studies mainly focus on images and thus are less useful for advertisement scenarios where rich auxiliary data are available. Here we propose a modality fusion-based neural network that evaluates the aesthetic preference of images with auxiliary information. Our method fully utilizes auxiliary data by introducing multi-step modality fusion using both conditional batch normalization-based low-level and attention-based high-level fusion mechanisms, inspired by the findings from statistical analyses on real advertisement data. Our approach achieved state-of-the-art performance on the AVA dataset, a widely used dataset for aesthetic assessment. Besides, the proposed method is evaluated on large-scale real-world advertisement image data with rich auxiliary attributes, providing promising preference prediction results. Through extensive experiments, we investigate how image and auxiliary information together influence click-through rate.
\end{abstract}

\section{Introduction}
\noindent 
Measuring image preference is both a fundamental and challenging problem because it is linked to the complex multimodal cognitive processes of humans visual perception~\cite{palmer2013visual,braun2013statistical}. Various studies have been conducted in the fields of cognitive science, computational photography/aesthetics, and neuromarketing to assess the emotional impressions and aesthetic qualities from images~\cite{murray2012ava,talebi2018nima}.

Image preference prediction can contribute to practical applications. Determining images for impressions: the number of times ads are displayed to customers, in online ads, is a prevalent application of image preference prediction. Predicting image preference for ads has two differences from the conventional aesthetic assessment. One is that an assessment model can use auxiliary information related to images, target subjects, and ad display policy. ad images are created by a designer for a purpose. It is a complex product that has undergone the creative processes of human designers. In particular, because the textual information is directly exposed to the content, the computational model must cope with both visual and language. The other difference is that an explicit metric can be used for evaluation, e.g., click-through rate (CTR).

Most existing methods on image assessment generally focused on an image as the input with respect to the quality and aesthetic. However, It has been reported that there exist other factors influencing on preference in addition to image quality, for example, colors~\cite{mehta2009blue,labrecque2012exciting}, brands, and verbal components~\cite{mitchell1986effect} were referred to affect traditional advertising. Furthermore, most CTR prediction studies heavily rely on user or meta information. There has been research on multimodal fusion with other information based on content~\cite{chen2016deep}. However, modality fusion can be more advanced to improve preference performance. Talebi and Milanfar~\cite{talebi2018nima} proposed a model to quantitatively evaluate human subjective aesthetic judgments about images. However, this approach might be ineffective for evaluating the advertising images due to not utilizing auxiliary information.

Here we propose a new approach to predict the preferences of users for determining which image is impressed to be advertised. For achieving this, we statistically analyzed the CTR for impressions, exposure events of online ads. Also, we explored the neural network structure inspired by the known knowledge and notions of existing advertisement professionals. i) We approached the CTR prediction problem as a regression problem rather than a classical classification. ii) Preprocessed as in image assessment tasks. iii) Inspired by statistical analysis and cognitive science, we propose a multi-step modality fusion network (M2FN). The term: multi-step refer to between images and auxiliary information, using conditional batch normalization (CBN)~\cite{de2017modulating} and a spatial attention mechanism for predicting CTR.


To evaluate our approach and model performance, we validated M2FN on two datasets. Our model is evaluated on impression CTR predictions for real-world ad images. AVA: an aesthetic assessment dataset similar to our content, was tested on M2FN for impression assessment. We achieved both image and impression assessments state-of-the-art results. With neural network visualization, we analyzed through various and sufficient experimental results in which area of the image influences the user's preference.

\begin{figure*}[ht]
		\includegraphics[width=\textwidth]{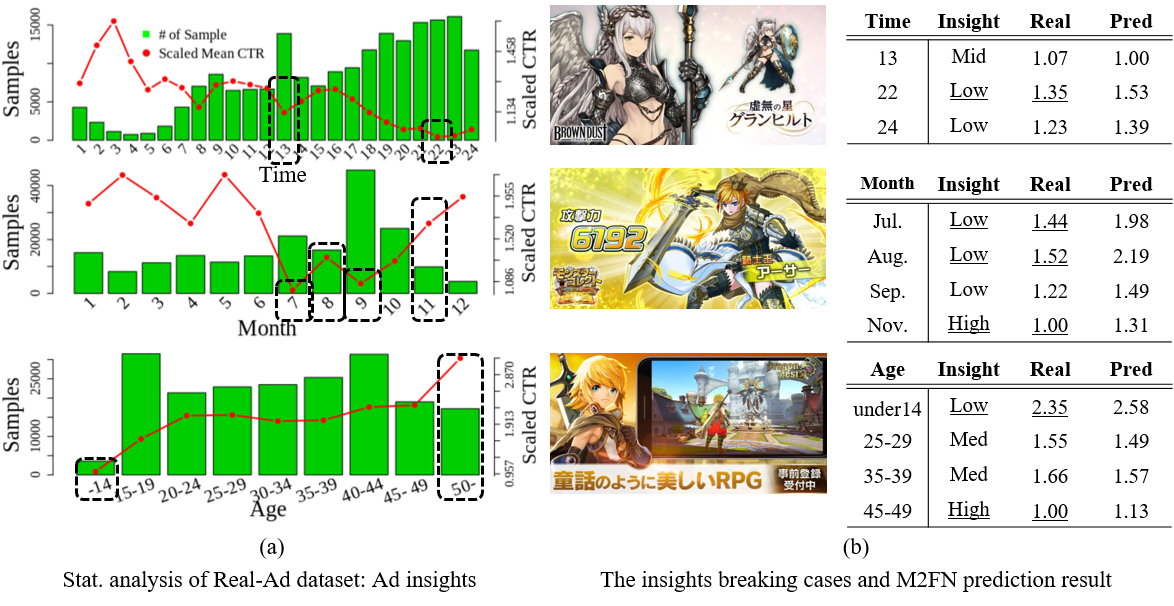}
		\caption{Advertisement insights and task introduction. (a) shows statistical analysis results of Real-Ad, a real-world online ad CTR dataset. It supports findings of existing literature and gives insights about ad impressions. One of the insights is that older users click ads more. (b) shows counter-evidence of insights derived from (a) and impression assessment results of our proposing model: M2FN. Although the table contents shown in (b) are different from the actual statistical analysis results in (a) (the ad insights), M2FN shows reliable performance. The CTR values are divided by the smallest CTR in each table due to confidentiality.}
		\label{myth_counter_cases}
\end{figure*}

\section{Related Work}
\textbf{Image Assessment: }Visual aesthetic assessment is associated with art, beauty, and personal preference. 
Driven by the importance of aesthetic assessment, aesthetic visual analysis (AVA) dataset, a large-scale image database accompanied by a variety of meta-data and rich annotations, was introduced for advance research~\cite{murray2012ava}. Using the AVA dataset, several studies have been carried out to predict aesthetic preference by reflecting the universally acceptable factors through learning~\cite{deng2017image,talebi2018nima,yu2018aesthetic}; although aesthetic preference depends on individual taste, some universal rules were reported such as the golden ratio, color harmonies, and the rule of thirds~\cite{datta2006studying,dhar2011high,ke2006design,luo2011content,luo2008photo}.
To obtain a higher correlation with human ratings, one of the papers~\cite{talebi2018nima} focused on the distribution of ratings instead of the aggregated mean scores using the earth mover's distance (EMD)~\cite{zhang2018unreasonable}. Most visual aesthetic assessment studies mainly focus on image contents rather than using addition textual metadata together. 

\noindent\textbf{Vision and Language: }Study on visual-linguistic representation learning has become popular thanks to the advent of Visual Question Answering (VQA) Challenge~\cite{iccv2015vqa}.
Generally, there are two key components in the learning methods.
1) Attention mechanism. 
Due to the nature of image or video, most methods reduce redundancy on spatio-temporal inputs based on other modalities, e.g., text~\cite{eccv2018kim}.
2) Multimodal fusion. 
Combining vision and language of different properties is not trivial. Various methods have been proposed from simple ones such as concatenation, element-wise addition to complex ones, e.g., compact bilinear pooling~\cite{emnlp2016Fukui}.
CBN is a fusion method that modulates visual prior by controlling batch normalization parameters in layers of visual representation model based on other inputs~\cite{de2017modulating}.
Our approach adopts ideas from the visual-linguistic representation learning methods in that we utilize auxiliary information by multi-step fusion based on both CBN and attention-based fusion mechanisms.

\noindent\textbf{Advertisement Preference Prediction: }
Recently, with the success of deep learning, neural network-based approaches have been proposed to predict CTRs, a widely used metric for evaluating ad performance. Compared to the conventional handcraft feature-based approaches, automatic and flexible feature learning directly from raw images has been introduced~\cite{mo2015image}. For more accurate prediction, the paper used ad information, such as detailed category and display position, in addition to raw images and confirmed performance improvements. Another CTR prediction method built an end-to-end deep learning architecture that learns representative features~\cite{chen2016deep}. They are from both raw images and other related information, such as ad zone, ad group, category, and user features. Even if these approaches use additional meta-information other than images, they have a room for improvement due to the use of simple concatenation-based modality fusion.

\section{Advertisement Images with Auxiliary Data}
\subsection{Task Definition and Data}
The problem addressed in this paper is to predict human preference for ad images with auxiliary information, which is formulated as a regression task. We define CTR as user preference on ad images. To solve this problem, we construct a large-scale dataset. It includes logs from ad displaying events represented with images, textual metadata, and their CTR value in a Japanese online ad service, called Real-Ad dataset. 
Therefore, ad images and their auxiliary information are used as input, and the CTR score is defined as the output value of regression.
{
\setlength{\tabcolsep}{2pt}
\ctable[
    caption = {Details of Real-Ad datasets. Instances of the raw dataset represents impressions per instance, and are different from instances of aggregated datasets. Instances of the raw dataset that have more than 100 and 500 impressions are aggregated to construct train and test datasets respectively.},
    label = {table:stat_real_ad_dataset},
    pos = h,
    doinside = \footnotesize
]{ccc rrr}{}{
    \toprule
    \multicolumn{3}{c}{Dataset} & \# instances & \# clicks & \# ad images  \\ 
    \midrule\midrule
    &&Raw(Game) & 500M & 20M & 3,747 \\ 
    \hline
    \multirow{4}{*}{Aggregated}&\multirow{2}{*}{100+}&Train & 353,510 & 800K & 3,045 \\
    &&Test & 173,248 & 330K & 1,436 \\
    \cmidrule{2-6}
    &\multirow{2}{*}{500+}&Train & 47,325 & 350K & 1,519 \\
    &&Test & 24,003 & 140K &  681 \\
    \bottomrule
}

}
\subsection{Real-Ad Data Construction}
An instance of raw CTR data is represented with one image and multiple auxiliary attributes. And it corresponds to one exposure event of ad content. The label of an instance is 1 or 0, denoting clicked or not-clicked. One instance is viewed many times with different times and target users.
All instances exposed with the same auxiliary information and ad image are aggregated to be the CTR score of the unique impression: 
\begin{equation}
{y_n} = \frac{{M_n^c}}{{{M_n}}},    
\end{equation}
\noindent where $M_n^c$ and $M_n$ denote the numbers of clicked and total ad displayed instances of the $n$-th unique exposure.

These preprocessed data are suitable for CTR prediction. 
The human preference for ad images may be represented as a distribution of real numbers with higher resolution. This facilitates comparative analysis with existing image assessment studies~\cite{talebi2018nima}. As a real-world task, it makes it possible to cope with label imbalance inevitable in CTR prediction. In the raw data, more than 99\% of the impressions are not clicked. The more frequent there is a conflicting label for the same instance, the worse the learning of the prediction models. Also, in real-world, these click data has an enormous scale, and thus those thousands of instances need to be compressed to increase expressiveness.

Following Table~\ref{table:stat_real_ad_dataset} shows the specification of the Real-Ad dataset we have constructed. The datasets appear to be very compressed compared to the raw data.

\subsection{Auxiliary Data}
In the existing image assessment studies, only images were used for scoring. But we went further from there: using additional auxiliary data. In general CTR studies, most input data consist of metadata such as user demographical data\textemdash gender and age, and ad exposure event-related data\textemdash date and time. We use all of these metadata, as well as additional data from a variety of sources. For example, the catchphrase of the ad is expected to contain the intent of the ad. To make computational models to learn this catchphrase, we used linguistic attributes in Real-Ad dataset. The dataset also has titles and descriptions that are commonly found in image datasets for linguistics information, but above all, OCR. If the ad image contains textual expressions such as typography, it is expected to be captured by a computational model trained with OCR auxiliary attribute. The details of used auxiliary attributes are explained in the following Datasets section. Linguistic auxiliary attributes are transformed into embedding vector by BERT~\cite{devlin2018bert}, and the rest are given with one-hot encoding vectors. By doing this, the various forms can be taken into account.

\subsection{Statistical Analysis on Real-Ad Data}
We investigate how ad images and their auxiliary attributes have influences on CTR by statistical analyses such as ANOVA and Logistic Regression Analysis. Through the statistical analysis, we could figure out the specific response of CTR along with each of the auxiliary attributes. Generally, these characteristic movements (ad insights) are strategically considered points by marketers to raise the impressions of ads. Among the impression-effective attributes, Figure~\ref{myth_counter_cases}-(a) indicates the ad insights of three attributes usually affect the performance of ads regardless of the category of the item being advertised. For example, Figure~\ref{myth_counter_cases}-(a)-the age bar plot shows those game ads have higher chances of being clicked when they are exposed to the older. In addition, we found that dominant color and linguistic attributes are also significant for ad images.  

On Top of that, the time and month bar plots in Figure \ref{myth_counter_cases}-(a) show how time-sequential attributes have an impact on CTR. In detail, ads published at dawn, morning, lunch, and day time recorded better result compare to other times. The beginning and the end of a week also showed higher CTR than other days of the week, shown in the supplementary materials. Lastly, the lowest CTRs occur in the third quarter of the year: July, August, and September, corresponding to the summer season.

On the other hand, there exist many cases which are not following these ad insights. Figure~\ref{myth_counter_cases}-(b) shows unusual cases found during statistical analysis of Real-Ad and the results of CTR prediction using M2FN. These analyses results show the preference prediction on ad images is very challenging. For example, it is expected that the CTR distribution will be different because of the same ad image and auxiliary attributes but different time attribute. This expecting phenomenon is similar to a covariate shift.

For addressing these issues, we need to design a model to effectively utilize auxiliary attributes and integrate them with ad images based on the found ad insights. 

\begin{figure}[t]
	\begin{center}
		\includegraphics[width=\columnwidth]{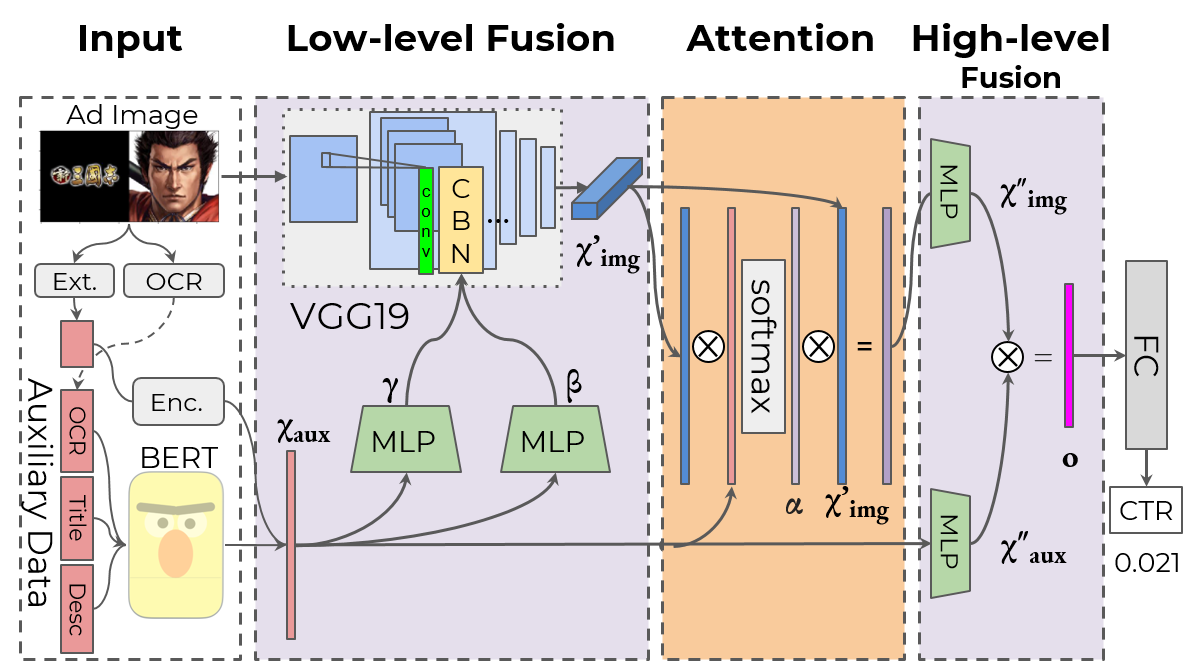}
		\caption{Overall structure of Multistep Modality Fusion Network (M2FN)}
		\label{overall_structure}
	\end{center}
\end{figure}

\section{M2FN: Multistep Modality Fusion Network}
We describe our model (M2FN) for predicting user preference from ad images and auxiliary information, inspired by the ad insights through Real-Ad data analysis. 
For fully utilizing these findings, M2FN consists of three main modality fusion steps. i) low-level fusion to deal with specific features such as dominant colors and auxiliary information. ii) spatial attention to address text expression location in ad images. iii) high-level fusion to consider abstracted visual features and auxiliary attributes such as demographics, time, and semantics of linguistic information.
 
 

\subsection{Low-level Fusion}
We conjecture that it can improve prediction performance to give each layer of the network a hint of the complexity of the ad image, auxiliary information. For achieving this, we employ CBN~\cite{de2017modulating}. It modulates the layers of the network instead of simple concatenation. CBN trains shallow neural networks which predict the scale factor parameters $\gamma$ and shift factor parameters $\beta$ as in batch normalization from the input data. Since the existing literature dealt with the VQA task, a question sentence is embedded using LSTM, and the question embedding $e_q$ was fed into the CBN module. Instead of text question, the auxiliary information is used as the input of the CBN. To be given to the CBN, categorical attributes are encoded with a one-hot vector, and text information is represented with a BERT-based embedding vector. 

CBN can be located at any layer of neural networks like Batch Normalization. In this paper, CBN was applied to fuse images and auxiliary information only after the first convolution of the early stage VGG-19, according to the best-performing experimental results.

\subsection{Attention Mechanism}
It is known that the location of textual information, as well as its expression in ad images, has a crucial influence on ad performance. To address this, we introduce an attention mechanism considering spatial relationships for modality fusion. 

In specific, image features and auxiliary embedding vectors are represented with an $N_b \times C \times W \times H$ and an $N_b \times  dim_{aux}$ tensors, where $N_b$, $C$, $W$, $H$ denote batch size, the number of channels, width, and height, respectively. The auxiliary embedding vectors are replicated to be a $N_b \times  dim_{aux} \times W \times H$ tensor. The auxiliary tensor is concatenated with the image feature tensor, to be a $N_b \times  (C+dim_{aux}) \times W \times H$  tensor. After that, it is fed into a fully-connected layer (MLP). The matrix obtained by softmax function the resulting vector $\{N_b, W \times H\}$ becomes the attention matrix. The attention matrix was multiplied with image features to achieve a soft-attention map.

\subsection{High-level Fusion}
The role of High-level fusion is to emphasize the effect of spatial relationships between visual feature and auxiliary information by integrating two modalities near output layer. Dissimilar to previous approaches, where a simple concatenation is used, matrix element-wise multiplication of the output of the attention mechanism and auxiliary information is performed as in~\cite{de2017modulating}. An affine transformation is performed to match the dimension size by linear layers for each of the image feature and auxiliary information. Then, the vectors from the activation function using hyperbolic tangent are element-wisely multiplied. There was a significant difference in regression performance with and without this fusion mechanism. It is shown in detail in the experiments section below.

\subsection{Loss Function}
Our model uses the impression weighted mean squared error as a loss function:
\begin{equation}
L = \frac{1}{N}\sum\nolimits_{n = 1}^N {{w_n} \cdot {{\left( {{{\hat y}_n} - {y_n}} \right)}^2}},
\end{equation}
\noindent where $w_n$, ${\hat y}_n$, and ${y_n}$ denote the impressed number, the predicted CTR, and the real CTR of the $n$-th sample. $N$ is the data size.
In addition, Kullback–Leibler divergence (KLD) is used as a loss function when the preference score of data is represented as a distribution form like AVA:
\begin{equation}
{L_{KLD}} = \frac{1}{N}\sum\nolimits_{n = 1}^N {p({y_n}) \cdot \log \frac{{p({y_n})}}{{p({{\hat y}_n})}}},
\end{equation}
\noindent where $p({y_n})$ and $p({{\hat y}_n})$ refer to the distributions of the real CTR and the CTR predicted by softmax function of the $n$-th sample, respectively.


\section{Datasets}

\subsection{Real-Ad Dataset} 
For evaluation, we use a large-scale dataset of click logs for ad image impressions, collected from a global online ad service in Japan during 2018. The total number of impressions is approximately 500 million, which include distinct 3,747 ad images. As constructing training data, we followed two steps such as aggregation, explained in the previous section, and attribute selection. 

\subsubsection{Attribute selection}
Raw Real-Ad contains about 40 attributes which were collected with click logs. Some attributes have weak effects on CTR, and there exist redundant attributes. To select attributes crucial for CTR prediction, we employed Analysis of Variance (ANOVA) and Logistic Regression Analysis under a significant level of 0.05 and thus selected nine attributes. The attributes selected for M2FN include \emph{gender, age, month, weekday, time, position (the position of ad in the displayed app page), category2 (mid-category of game, e.g. casual, hard-core, etc), category3 (low-category of game, e.g. role-playing, action, gambling, etc), and dominant color}. More details can be found in the supplementary material.

\subsubsection{Attribute preprocessing}
Dominant color and linguistical auxiliary attributes are required to be preprocessed for learning, including title, description, and embedded textual expression of ad images (OCR). The dominant color of an ad image is represented as an element of the predefined color set, including ten colors. For achieving this, we use $K$-means clustering with the minimum covariance determinant (MCD) distance for extracting intermediate dominant color from the image. After obtaining the intermediate color, we mapped it to one of the predefined dominant colors.

Also, we supplemented linguistical auxiliary attributes such as title, description, and OCR result. The title and description were collected from the content introduction, such as smartphone app markets. OCR result is obtained by the open API\footnote{https://github.com/clovaai/CRAFT-pytorch}~\cite{baek2019character,baek2019STRcomparisons} to identify the letters in the ad images. Unique sentences collected for each of these attributes: The numbers of sentences are 1,583, 2,695, and 2,695, respectively. We embedded them into vectors using BERT\footnote{https://github.com/huggingface/pytorch-transformers}~\cite{devlin2018bert}.

While grouping and summing CTRs, we only considered instances which have more than 100 and 500 impressions for the feasibility of CTR value. We merged levels of the attribute, which has less than 50,000 impressions to the closest level for coping with being biased in the dataset.

In Real-Ad dataset, the total dimension of auxiliary attributes is 2,383, and the linguistics are majority: 2,304 (768 dimensions for each).

\subsection{Benchmark dataset: AVA Dataset}
M2FN can contribute to a conventional image assessment task. We evaluate our model on AVA~\cite{murray2012ava}, which is an image dataset designed for aesthetic preference study. This dataset consists of images and their annotations-aesthetic, semantic, and photographic style annotations. The photographic style annotations are comprised of 14 styles (complementary colors, duotones, HDR, etc.) to represent the camera settings. However, these annotations are not used in this paper due to too many no-annotation cases. 

Semantic annotations are textual tag data such as \textit{nature, black and white, landscape, still life, macro, animal}, etc. The number of images is approximately 200k, and each image has at least one tag. Tag set size is 67 by adding no-tag cases for image-only CTR prediction. A text tag is represented as an embedding vector by BERT. These embedding vectors with 768 dimensions were used for the experiment as linguistical auxiliary attributes.

Aesthetic annotations are rating data per image from hundreds of amateur and professional photographers. These data are characterized by the histogram distribution ranging from 1 to 10, which are used as score labels. The dataset was divided into the training, and the test sets with an 8:2 ratio for benchmark experiments.


\section{Experimental Results} 
Explanation of terms: The resulting outputs may be in the form of distribution (10 buckets) or a scalar value (a score) depending on the dataset. Performance evaluation is based on ranking. Spearman rank correlation (SPRC) and linear correlation coefficient (LCC) are computed to rank the output scores and compare them to the ranking of the ground truth. If the output score is in the form of distribution, SPRC and LCC are calculated and examined for both mean and standard deviation. 

The benchmark and Real-Ad datasets have distinct hyperparameter settings for M2FN. Among the four major modules, CBN (low-level fusion), attention, and high-level fusion should determine the hidden layer size of the multi-layer perceptron (MLP) for embedding representations inside the module. When training AVA datasets which have relatively small auxiliary information, the hyperparameters of the modules are 64, 512, and 512, respectively. When training the Real-Ad dataset, we decided to 256, 512, and 1,024. This decision was made based on experiments. The batch size of the dataset was 128, and five P40 GPUs were used for training. All models are trained for 100 epochs. All the experiments are implemented and performed based on NAVER Smart Machine Learning (NSML) platform~\cite{kim2018nsml,sung2017nsml}.

{
\setlength{\tabcolsep}{2.2pt}
\ctable[
    caption = {Performance comparison on Real-Ad. m and std. in the header row respectively represent mean and standard deviation.},
    label = {table:model_comparison_real_ad},
    pos = h,
    doinside = \footnotesize
]{cclccccc}{}{
    \toprule
\multicolumn{3}{c}{Models}                      & SPRC(m) & LCC(m) & SPRC(std.) & LCC(std.) \\ 
               \midrule\midrule
\multirow{5}{*}{100+} & \multirow{3}{*}{Dist.}   & NIMA          & 0.110   & 0.121  & 0.146      & 0.142     \\
                      &                           & NIMA*         & 0.289   & 0.290  & 0.139      & 0.143     \\
                      &                           & \textbf{M2FN} & 0.344   & 0.367  & 0.172      & 0.175     \\ 
                      \cmidrule{2-7}
                      & \multirow{2}{*}{Regr.} & NIMA          & 0.325   & 0.343  & -          & -         \\
                      &                           & \textbf{M2FN} & \textbf{0.384}   & \textbf{0.381}  & -          & -         \\ 
                      \midrule
\multirow{5}{*}{500+} & \multirow{3}{*}{Dist.}   & NIMA          & 0.308   & 0.249  & 0.165      & 0.166     \\
                      &                           & NIMA*         & 0.453   & 0.448  & 0.190      & 0.190     \\
                      &                           & \textbf{M2FN} & 0.484   & 0.478  & 0.216      & 0.190     \\
                      \cmidrule{2-7}
                      & \multirow{2}{*}{Regr.}  & NIMA          & 0.501   & 0.451  & -          & -         \\
                      &                           & \textbf{M2FN} & \textbf{0.561}   & \textbf{0.530}   & -          & -         \\
    \bottomrule
}
}

{
\setlength{\tabcolsep}{2pt}
\ctable[
    caption = {Performance comparison on Real-Ad. m and std. in the header row respectively represent mean and standard deviation.},
    label = {table:model_comparision_AVA},
    pos = h,
    doinside = \footnotesize
]{lcccc}{}{
    \toprule
    \multicolumn{1}{c}{Models} & SPRC(m) & LCC(m) & SPRC(std.) & LCC(std.) \\ 
    \midrule\midrule
    NIMA(Inception-v2) & 0.612 & 0.636 & 0.218 & 0.233 \\
    M1FN(VGG19-cat) & 0.572 & 0.581 & 0.198 & 0.197 \\ \addlinespace
    \textbf{M2FN}& \textbf{0.630} & \textbf{0.640} & \textbf{0.310} & \textbf{0.322} \\
    \bottomrule
}
}

Table~\ref{table:model_comparison_real_ad} compares the CTR prediction performance on the Real-Ad dataset. 
Unlike AVA, which includes rating counts in the form of ten buckets for each image, Real-Ad only provides a mean of CTRs. Since our proposed method requires training on a CTR value, the CTR distributions are approximated by a log-normal distribution which resembles CTRs. When converted to a distribution, it is represented as ``dist."; otherwise, it is described in Table~\ref{table:model_comparison_real_ad} as ``regr.". Previously, the best performing model (NIMA) on the benchmark dataset used EMD as a loss function. The star (*) means that the KLD is applied as a loss function instead of the EMD (for ``regr." cases, the impression weighted MSE is used as mentioned above section). 

Regardless of the number of impressions (100+ or 500+), it can be seen that our method performs better. In our dataset, KLD appears to be a better choice as a loss function than the EMD.

In Table~\ref{table:model_comparision_AVA}, we compare our models to NIMA~\cite{deng2017image} on the AVA dataset for verifying that our model can be applied to image assessments., with a quantitative comparison between NIMA (Inception-v2), the previous SOTA model, and our M2FN.

VGG19-cat in Table~\ref{table:model_comparision_AVA} describes the preliminary method we have implemented for fusing the auxiliary information: ``-cat" stands for simple concatenating of both the auxiliary and image. As a result, the model adopts a simple fusion (concatenating) deteriorates performance. M2FN, which is using multi-step modality fusion performed best in the benchmark dataset. 

{
\setlength{\tabcolsep}{5.5pt}
\ctable[
    caption = {M2FN ablation study results on Real-Ad},
    label = {table:ablation_module_real_ad},
    pos = h,
    doinside = \footnotesize
]{cccc|cc|cc}{}{
\toprule
\multicolumn{4}{c}{Module}      & \multicolumn{2}{c}{500+} & \multicolumn{2}{c}{100+} \\ 
\midrule
Aux&Low&Att&High                         & SPRC             & LCC                              & SPRC                             & LCC \\ 
\midrule\midrule
$\times$&$\times$&$\times$&$\times$          & 0.456              &   0.435                      &  0.315                             &  0.342\\
\midrule
O&$\times$&$\times$&$\times$              & 0.437    &  0.412          & 0.297                &  0.320\\ 
O&O&$\times$&$\times$               & {0.496}           &   {0.498}      &  {0.367}                &  {0.371}\\
\midrule
O&$\times$&O&$\times$               & 0.450                    &  0.463                          & 0.341                        &  0.325\\
O&$\times$&$\times$&O               & 0.475                 &  0.467                          & 0.321                          &  0.293\\
O&O&O&$\times$                     & 0.506                &   0.480                        &  0.356                            &  0.334\\
O&O&$\times$&O                    & 0.554                & 0.528                          &  0.361                           &  0.333\\ 
\midrule
O&O&O&O                         & \textbf{0.561}         & \textbf{0.530}               & \textbf{0.384}                 &   \textbf{0.381}\\
\bottomrule
}
}

Table~\ref{table:ablation_module_real_ad} compares the performance changes according to the presence or absence of four major modules. The four primary modules are auxiliary data, low-level fusion, attention, and high-level fusion. In the four consecutive columns, ``O" stands for activating, and ``$\times$" stands for deactivating. ``$\times \times \times \times $" is a model trained only images on vanilla VGG-19, the basis of M2FN. In the case of ``O $\times\times\times$" where the auxiliary data is activated, but the CBN is deactivated, it means that the image and auxiliary information are learned by concatenating before the last fully connected layer.

As can be seen in the results of the second and third rows indicating the presence and absence of low-level fusion, we can see that CBN boosts performance. It also supports the health of M2FN once again in the results of the fourth row. It indicates the presence of high-level fusion. The best performance is achieved when all four modules are combined. The four modules integrated result proved that our approach is suitable for evaluating impressions.

{
\begin{table}[h]
\centering 
\caption{M2FN ablation study of auxiliary information using Real-Ad. The attributes start with upper cases represent consist of multiple attributes.}
\begin{threeparttable}
\begin{tabular}{c ccc cc c c c}
    \toprule
    &\multicolumn{3}{c}{Used auxiliary attributes}                &&   \multicolumn{1}{c}{SPRC}  &&  \multicolumn{1}{c}{LCC}&\\  
    \midrule\midrule
    &\multicolumn{1}{l}{None}&&                             &&   \multicolumn{1}{c}{0.456}       &&   \multicolumn{1}{c}{0.435}&\\ 
    \midrule
    &\multicolumn{1}{l}{ALL\tnote{1}}&&                             &&   \multicolumn{1}{c}{\textbf{0.561}}       &&   \multicolumn{1}{c}{\textbf{0.530}}&\\ 
    &&\multicolumn{1}{l}{$-$ date}   &&&   \multicolumn{1}{c}{\underline{0.454}}       &&   \multicolumn{1}{c}{\underline{0.413}}&\\  
    &&\multicolumn{1}{l}{$-$ description}   &&&   \multicolumn{1}{c}{\underline{0.467}}       &&   \multicolumn{1}{c}{\underline{0.453}}&\\  
    &&\multicolumn{1}{l}{$-$ color}   &&&   \multicolumn{1}{c}{\underline{0.503}}       &&   \multicolumn{1}{c}{\underline{0.470}}&\\  
    &&\multicolumn{1}{l}{$-$ (User+Text)\tnote{2}}   &&&   \multicolumn{1}{c}{0.513}        &&   \multicolumn{1}{c}{0.507}&\\  
    &&\multicolumn{1}{l}{$-$ ocr}   &&&   \multicolumn{1}{c}{0.515}       &&   \multicolumn{1}{c}{0.491}&\\  
    &&\multicolumn{1}{l}{$-$ gender} &&&   \multicolumn{1}{c}{0.532}       &&   \multicolumn{1}{c}{0.470}&\\  
    &&\multicolumn{1}{l}{$-$ age}   &&&   \multicolumn{1}{c}{0.535}       &&   \multicolumn{1}{c}{0.486}&\\  
    &&\multicolumn{1}{l}{$-$ User\tnote{3}}  &&&   \multicolumn{1}{c}{0.539}       &&   \multicolumn{1}{c}{0.514}&\\  
    
    \midrule\midrule
    &\multicolumn{1}{l}{Text}&&               &&   \multicolumn{1}{c}{0.422}       &&   \multicolumn{1}{c}{0.369}&\\
    &&\multicolumn{1}{l}{$-$ (title $+$ ocr)}&&                       &   \multicolumn{1}{c}{0.453}       &&   \multicolumn{1}{c}{0.411}&\\
    \midrule
    &\multicolumn{1}{l}{User}&&  &&   \multicolumn{1}{c}{0.437}       &&   \multicolumn{1}{c}{0.397}\\
    \bottomrule
\end{tabular}

\begin{tablenotes}\footnotesize
    \item[1] ALL aux: Text, User, date, time, position, categories, color
    \item[2] remove Text (title, desc, ocr) with User(gender, age) auxes.
    \item[3] remove User (gender, age) from ALL auxes.
    
\end{tablenotes}
\end{threeparttable}

\label{table:ablation_aux_real_ad}
\end{table}
}

\begin{figure*}[ht]
		\includegraphics[width=\textwidth]{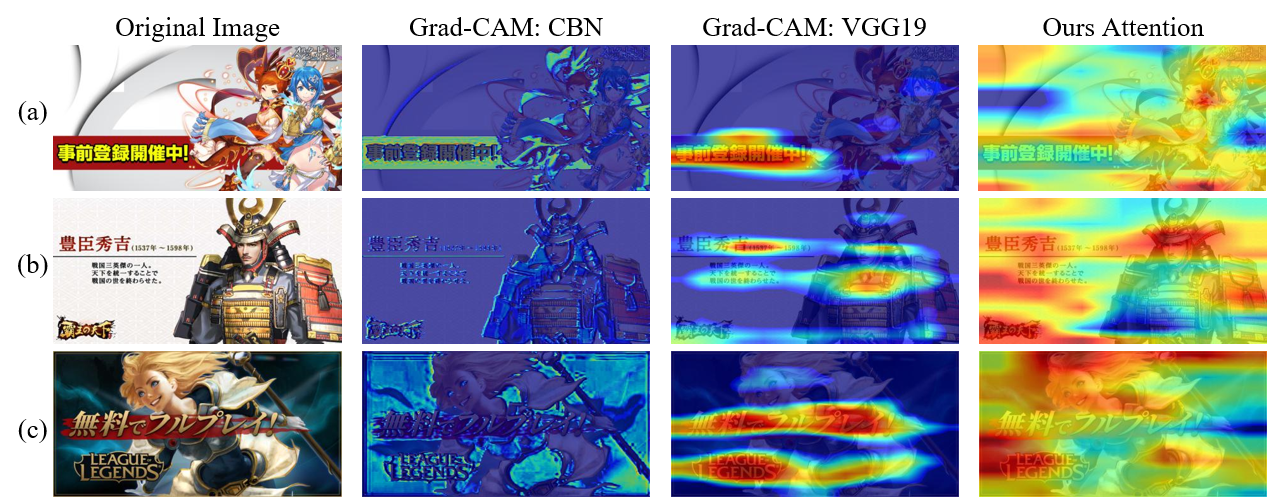}
		\caption{Visualization results using Grad-CAM. Each column displays heatmaps that depict weight vectors of specified layers. The red and blue colors respectively represent the highest value and the lowest value. Best viewed in color.}
		\label{gradcam_results}
\end{figure*}

The results in Table~\ref{table:ablation_aux_real_ad} are ablation study comparing performance changes with or without auxiliary attributes in the Real-Ad dataset. Table~\ref{table:ablation_aux_real_ad} shows only a part, and the entire experimental results can be found in supplementary materials. Because we conjectured content-related attributes (date, time, position, categories, and color) are more information than others, e.g., user-related attributes, for evaluating impressions. We expected that excellent performance would be achieved even if we used only content-related attributes among auxiliary attributes. As we expected, the result proves our conjecture.

We sorted the performance in Table~\ref{table:ablation_aux_real_ad} in ascending order. If an attribute is excluded and the performance drop is significant, it means that the attribute is essential. As a result, we can see the following: i) The time-sequential attribute (date) is vital. ii) According to the bottom of the table (-(title + ocr)), description has little effect when used alone. But it is having a significant impact when it is integrated with other auxes. iii) As the previous studies asserted, the color attribute proved itself essential for advertising content in the table again. iv) The user attribute is not crucial, even though it heavily used in previous studies. v) However, when the user attribute is combined with other (text aux in the table), the performance degradation seems to be substantial. It is assumed that the user attribute has its potential when integrated.

The above five findings indicate that there is a complex action between auxiliary attributes. Therefore our hypothesizes were correct.



\subsection{Further Analysis}
Statistical and experimental results provide some insights, which can be used for assessing the attractiveness of ad images induced by a human designer. 
Based on the statistical analysis of the Real-Ad, we reaffirmed previous studies: Ad insights of single or combination of the attributes such as time, month, date, and age. As listed in Table~\ref{table:ablation_aux_real_ad}, we show that the four attributes are very influential. The model was robust even under exceptional circumstances that did not follow the insights. Detailed case studies can be found in the supplementary material. 
Also, with the combination of the attributes, the fused representation in M2FN fires like an activation function in neurons: e.g., united they fires, divided they don't. All of this would be very interesting from the perspective of cognitive science and marketing research. 

In Fig~\ref{gradcam_results}, each layer of M2FN was leveraged using Grad-CAM~\cite{selvaraju2017grad} to visualize the operation of each layer in the model. Grad-CAM is a tool that allows us to see which part of an image the neural network sees and makes a decision on a particular label. It facilitates the layer to understand the importance of each neuron using the gradient information. In Fig~\ref{gradcam_results}, heatmaps are drawn, which is shown in red as a vital area. Interestingly, M2FN seems to be heavily influenced by the text on the image after the extraction of visual features (third column in the figure). As a result, in the attention map (fourth column), a visualization result also shows having a salience in part with characters and typography was obtained. Analyzing these results, our model learned well where the linguistic elements that humans consider necessary in advertising images are located. 
The model gives a hint about human's visual-spatial saliency toward the ad. This proves the results of existing literary and cognitive or marketing science studies, and at the same time, M2FN has achieved preliminary success in learning the human aesthetics toward advertising.

\section{Conclusion}
In this paper, we propose a model for predicting user preference of ad images. We collect a large-scale dataset, including ad images and auxiliary data, called Real-Ad dataset. Then, we statistically explore Real-Ad dataset, focusing on the influence of images and auxiliary attributes on human preference represented as CTR. Inspired by ad insights found from the analyses, we design a new multi-step modality fusion network (M2FN). M2FN is to effectively integrate ad images and their auxiliary attributes to predict CTR. We evaluate M2FN on Real-Ad dataset. Besides, we validate our method on a benchmark image assessment dataset, AVA dataset for verifying whether our approach can be applied to conventional image assessment. M2FN achieved better performance on both datasets compared to the previous state-of-the model. With extensive ablation study, we investigate how each modality fusion works and which auxiliary attributes largely influences user preference.

As further works, we will implement a generative model that uses M2FN as a discriminator. We consider to generate ad images that can achieve higher CTR.
\bibliography{main_paper.bib}
\bibliographystyle{aaai}

\end{document}